\documentclass[conference]{IEEEtran}
\IEEEoverridecommandlockouts
\usepackage{cite}
\usepackage{amsmath,amssymb,amsfonts}
\usepackage{algorithmic}
\usepackage{graphicx}
\usepackage{textcomp}
\usepackage{xcolor}
\usepackage{subcaption}
\def\BibTeX{{\rm B\kern-.05em{\sc i\kern-.025em b}\kern-.08em
    T\kern-.1667em\lower.7ex\hbox{E}\kern-.125emX}}

\usepackage[compact]{titlesec}
\titlespacing{\section}{0pt}{2ex}{1ex}
\titlespacing{\subsection}{0pt}{1ex}{0.5ex}
\titlespacing{\subsubsection}{0pt}{0.5ex}{0ex}

\usepackage{enumitem}
\setlist{nosep} % Removes vertical space between list items

\usepackage[font=small,labelfont=bf,skip=2pt]{caption}

\begin{document}

\title{Food Portion Estimation: From Pixels to Calories\\

}

\author{Gautham Vinod
\and
Fengqing Zhu}

% \and
% \and
% \and
% \and
% \and

\maketitle

\begin{abstract}
Reliance on images for dietary assessment is an important strategy to accurately and conveniently monitor an individual's health, making it a vital mechanism in the prevention and care of chronic diseases and obesity. However, image based dietary assessment suffers from estimating the three dimensional size of food from 2D image inputs. Many strategies have been devised to overcome this critical limitation such as the use of auxiliary inputs like depth maps, multi-view inputs, or model based approaches such as template matching. Deep learning also helps bridge the gap by either using monocular images or combinations of the image and the auxillary inputs to precisely predict the output portion from the image input. In this paper, we explore the different strategies employed for accurate portion estimation. 
\end{abstract}


\section{Introduction}
\begin{figure*}
    \centering
    \includegraphics[width=\linewidth]{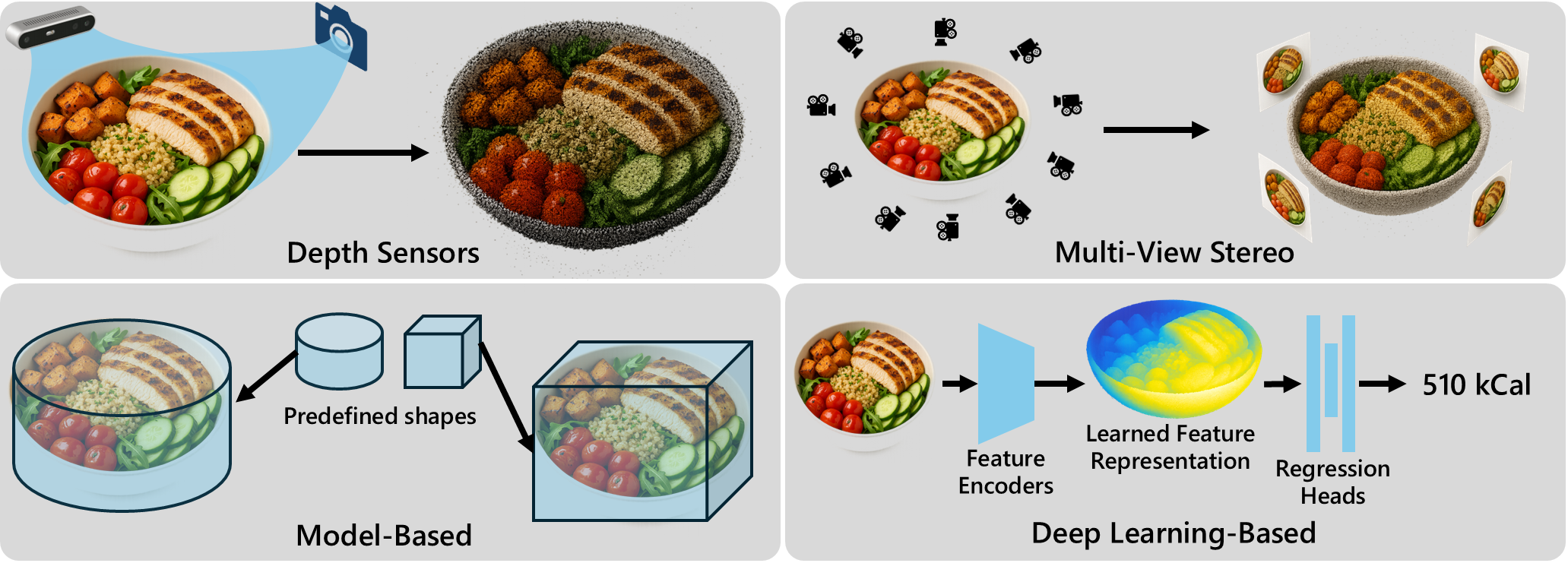}
    \caption{\textbf{Traditional and Deep Learning Methods}. Traditional methods of extrapolating missing 3D information from 2D images rely on depth-based (specialized hardware), multi-view stereo, or model-based methods. Deep learning tries to bridge the gap by using auxiliary inputs or learning to map the input image to the auxiliary input and using combinations of them for portion regression.}
    \label{fig:overview}
    \vspace{-0.1cm}
\end{figure*}

Dietary assessment is an essential tool in reducing the risk of chronic diseases such as cardiovascular issues, obesity, and diabetes, which are often linked to unhealthy dietary patterns~\cite{wang2023optimal}. 
However, traditional dietary assessment relies heavily on self-reporting methods, such as the 24-hour recall~\cite{baranowski201224} and food frequency questionnaires. In these procedures, individuals are asked to self-report their food intake over a preceding period to registered dietitians. This process is not only burdensome for the individual but also prone to significant errors; studies have consistently shown that participants frequently under-report consumption due to bias~\cite{karvetti1985validity, subar2015traditional}.

Image-Based Dietary Assessment (IBDA) aims to automate this process by allowing users to simply capture images of their eating occasions, thereby reducing user effort and reporting bias~\cite{Shao2023VoxelReconstruction}. A standard Image-Based Dietary Assessment pipeline typically consists of food detection, classification, and portion estimation. While traditional classification (identifying ``what'' is on the plate) has reached maturity due to large-scale datasets, food data remains particularly challenging due to its inter-class similarity (pasta sauce vs jelly) and intra-class variance (different types of burgers or sandwiches). More importantly, quantification (estimating ``how much'') remains a persistent bottleneck.

The core challenge in portion estimation is the \textit{scale ambiguity problem}. When a 3D real-world scene is projected onto a 2D image plane, depth information is lost. Without a known reference, a small pizza close to the camera is geometrically indistinguishable from a large pizza placed further away. Consequently, portion estimation from a single image is mathematically an ill-posed problem, as infinite 3D volumes can yield the exact same 2D projection~\cite{Vinod20243DFoodPortion}.

To resolve this ambiguity, early approaches relied on explicit geometric recovery. These methods often required \textit{fiducial markers} (e.g., a coin or checkerboard patterns of known size) placed in the scene to provide a physical reference for scale~\cite{jia2014accuracy}. While effective, this imposes an additional burden on the user, reducing the usability of the system. Subsequent research explored the use of specialized hardware, such as RGB-D sensors (e.g., Microsoft Kinect) or smartphone LiDAR, to capture direct depth maps~\cite{lo2019image, myers2015im2calories, Makhsous2019}. However, such sensors are not universally available on all consumer devices and often struggle with the fine-grained texture required to distinguish food density. Alternatively, multi-view stereo (MVS) techniques attempted to reconstruct 3D meshes from multiple images of the meal~\cite{dehais2016two}, but these require rigid scenes and significant user cooperation to capture 360-degree views.

Recent advancements in Deep Learning have shifted the focus from explicit geometric reconstruction to \textit{monocular inference}. Modern architectures leverage vast datasets to ``learn'' depth priors, enabling the prediction of volume and energy directly from single RGB images without physical references~\cite{myers2015im2calories, fang2021single, Shao2023VoxelReconstruction, vinod2022image}. These methods utilize techniques ranging from monocular depth estimation to implicit volumetric representations, such as Neural Radiance Fields (NeRFs)~\cite{mildenhall2020nerf}, to fill in the missing 3D structure.

This paper reviews the challenges and advancements in food portion estimation. We structure our analysis by first discussing geometric methods that recover 3D data via sensors or multi-view constraints. We then examine how deep learning has leveraged these inputs to enable monocular prediction, discussing the trade-offs between accuracy, user burden, and computational complexity.
\section{Geometric Size Estimation: Recovering 3D Data}

The fundamental challenge in portion estimation is the recovery of metric 3D structure from 2D projections. To estimate the volume $V$ of a food item (or other physical quantities related to size, such as weight), a system must resolve two unknowns: the absolute physical scale of the scene (metric depth) and the shape geometry of the object. We categorize geometric approaches into Depth Sensing, Multi-view Stereo, and Model-based, as shown in Figure~\ref{fig:overview}.

\subsection{Specialized Depth Sensors}
Early IBDA systems bypassed the ill-posed nature of monocular estimation by utilizing hardware that captures explicit depth maps, $D(u,v)$, alongside RGB data~\cite{thames2021nutrition5k, myers2015im2calories, lo2019image, raju2022foodcam, shang2011dietary}.

Structured Light or RGB-D: sensors, such as the Microsoft Kinect or Intel RealSense, project a known pattern of infrared light (e.g., speckles or grids) onto the scene. The deformation of this pattern on object surfaces allows for the calculation of depth via triangulation.

While highly accurate for volume integration, these systems suffer from critical usability limitations: they often fail on reflective or transparent surfaces (e.g., soup, metallic cutlery), perform poorly in outdoor lighting due to IR interference, and have low accuracy for smaller objects (such as food). Furthermore, the requirement for bulky, external hardware renders them impractical for daily, in-the-wild dietary tracking.

Mobile LiDAR and Time-of-Flight (ToF) scanners in consumer smartphones (e.g., iPhone Pro series) have sparked interest in direct depth capture. ToF sensors measure the round-trip time of light pulses to generate a depth map. However, the spatial resolution of mobile LiDAR is often insufficient for dietary assessment. Small, granular food items (e.g., rice grains, nuts) are frequently smoothed over, resulting in significant volume estimation errors due to the ``convex hull'' effect, where empty spaces between particles are counted as solid mass~\cite{keller2024nutritionverse, dinic2017eatar}.

\subsection{Multi-View Stereo (MVS)}

\noindent\textbf{Structure from Motion (SfM):} By capturing a video or a series of static images around a dining plate, SfM algorithms (e.g., COLMAP)~\cite{schoenberger2016mvs, schoenberger2016sfm, rahman2012foodvolume} simultaneously estimate the camera poses and generate a point cloud of the scene. While effective for rigid objects~\cite{he2024MetafoodChallenge}, SfM assumes a static scene; any movement of the food (e.g., steam, shifting contents) during capture introduces noise. Scale is often missing unless the intrinsic camera parameters are known.

\noindent\textbf{Stereo Matching:} Utilizing rectified stereo pairs~\cite{Dehais2017TwoViewFoodReconstruction}, depth $Z$ can be computed via disparity $d$, which represents the displacement of a point between the left and right views. The relationship is governed by the equation:
\begin{equation}
    Z = \frac{f \cdot B}{d}
\end{equation}
where $f$ is the focal length and $B$ is the baseline distance between cameras.

 However, the primary bottleneck for MVS in dietary assessment is user burden. Requiring a user to perform a 360\textdegree scanning motion or capture multiple stable angles is cumbersome. Furthermore, MVS struggles with the \textit{occlusion problem}: parts of the food touching the plate or obscured by other items cannot be reconstructed, necessitating interpolation that introduces volume error.

\subsection{Model-Based and Template Matching}
To mitigate the need for dense reconstruction, model-based approaches fit observations to known shape priors, effectively reducing the problem to a parameter estimation task.

\noindent\textbf{Geometric Primitives:} Foods with consistent structures are approximated using simple Euclidean shapes. For example, drinks are modeled as cylinders, bowls as semi-spheres, and cakes as triangular prisms. Volume is calculated analytically (e.g., $V = \pi r^2 h$ for a cylinder). While computationally efficient, this method generalizes poorly to non-rigid or amorphous foods (e.g., salads, curries) that do not conform to standard geometric primitives~\cite{xu2013model, fang2016comparison}.

\noindent\textbf{3D Template Scaling:} This method utilizes a library of high-fidelity 3D meshes (e.g., a scanned banana or apple model). The system projects the 3D template onto the 2D image plane and optimizes the pose parameters $\mathbf{R}, \mathbf{t}$ and a global scale factor $s$ to minimize the alignment error between the projection and the segmentation mask~\cite{Vinod20243DFoodPortion}.

Here, the rigidity of templates is a significant limitation. A generic pizza slice template cannot adapt to a specific instance that has a bite taken out or a folded crust. These discrepancies between the ideal template and the real-world deformation lead to substantial volumetric error. Further, a need for a known physical anchor reduces the use case of such methods.
\section{Deep Learning: The Monocular Shift}

The most significant advancement in recent years is the transition from explicit geometric recovery to \textit{monocular inference}. Deep Neural Networks (DNNs) have demonstrated the ability to ``infer'' missing 3D information from a single RGB image by learning priors from large datasets~\cite{myers2015im2calories}. This shift allows for portion estimation, sometimes from monocular inputs.

\subsection{Monocular Depth Prediction}
Instead of relying on a physical sensor, this approach treats depth estimation as a dense regression problem. A neural network learns a mapping function $f_\theta: I \rightarrow \hat{D}$, predicting a per-pixel depth map $\hat{D}$ from a single RGB image $I$~\cite{ma2024mfp3d, myers2015im2calories}.
Here, an encoder-decoder architecture is typically employed where the encoder extracts high-level semantic features (identifying the object as ``food''), while the decoder progressively upsamples these features to regress the depth value for each pixel $(u, v)$~\cite{Shao2023VoxelReconstruction}.

Once the depth map $\hat{D}$ is obtained, it is back-projected into 3D space using the camera intrinsic matrix $K$. The system generates a point cloud or voxel grid, and volume is computed by integrating the voxels.

Recent methods like DPF-Nutrition~\cite{han2023dpf} utilize \textit{cross-modal fusion}. Rather than treating depth and RGB features in isolation, these networks fuse semantic features (texture, category) with geometric features (shape, depth) at multiple scales, significantly reducing the error in volume estimation compared to depth-only baselines.

\subsection{Direct Energy Regression}
Another bypasses the intermediate 3D reconstruction entirely. These ``end-to-end'' architectures learn a direct mapping from image pixels to nutritional content: $I \rightarrow \mathbb{R}^N$, where output represents mass, volume, kilocalories, or other macronutrients~\cite{thames2021nutrition5k}.

Standard backbone networks are modified by replacing the final classification layer with a regression head. The network minimizes a loss function between the predicted calorie count and the ground truth~\cite{sanatbyek2025fpb}.

\subsection{Implicit Representations}
The cutting edge of food computer vision has adopted Neural Radiance Fields (NeRFs) to overcome the limitations of discrete meshes~\cite{mildenhall2020nerf}.

Unlike explicit point clouds, NeRFs represent the food scene as a continuous volumetric function $F_\Theta : (\mathbf{x}, \mathbf{d}) \rightarrow (\mathbf{c}, \sigma)$, mapping a spatial coordinate $\mathbf{x}$ and viewing direction $\mathbf{d}$ to a color $\mathbf{c}$ and density $\sigma$.

Methods such as those benchmarked in NutritionVerse~\cite{keller2024nutritionverse} represent a significant leap in quality. They can synthesize novel views and extract highly accurate volumetric meshes from very few input images (few-shot learning). Crucially, NeRFs handle semi-transparent and complex objects (e.g., soups, amorphous purees) significantly better than traditional Multi-View Stereo, which often fails to match features in texture-less or fluid regions~\cite{vinod2026sizematters}.

However, these various methods face three fundamental challenges: Scale ambiguity, Occlusion problems, and Density estimation.

% \subsection{Key Challenges}
% Despite these algorithmic strides, three fundamental challenges prevent fully automated, high-precision estimation:

% \begin{enumerate}
%     \item \textbf{Scale Ambiguity:} A monocular image lacks absolute metric scale. A generic DNN cannot distinguish between a small plate and a large platter without context. Current robust solutions still require a \textit{fiducial marker} (e.g., a standardized checkerboard, coin, or credit card) visible in the frame to establish a pixel-to-millimeter ratio~\cite{jia2014accuracy}.
%     \item \textbf{Occlusion:} Visual methods are limited to the visible surface. The ``hidden'' geometry—such as the bottom of a burger or the ingredients inside a burrito—must be inferred. DL models rely on learned priors to guess this hidden volume, often leading to over-smoothing or under-estimation in complex, piled food items~\cite{fang2016comparison}.
%     \item \textbf{Density Estimation:} The mapping from Volume ($cm^3$) to Mass ($g$) is non-linear and material-dependent. A fluffy cake and a dense brownie may share identical volumes but differ vastly in caloric content. Bridging this gap requires \textit{visual-tactile learning} or fine-grained material recognition to infer density from surface texture.
% \end{enumerate}
\section{Challenges and Future Directions}

While the shift from active sensors to monocular deep learning has improved usability, the problem of accurate portion estimation remains partially unsolved. We identify three critical bottlenecks and propose directions for future research.

\subsection{The Scale and Reference Problem}
The most persistent challenge in monocular estimation is \textit{scale ambiguity}. Without a known reference, the mapping from pixel coordinates $(u,v)$ to world coordinates $(X,Y,Z)$ is indeterminate. Current solutions rely on fiducial markers (e.g., a Checkerboard or credit card)~\cite{jia2014accuracy}, which disrupts the user experience. Deep learning strategies are dependent on the distribution of the training data often failing to accurately predict scale for in-the-wild input images.

Currently, research is moving toward \textit{marker-less scale estimation} by exploiting environmental priors. This includes dynamic reference scales or learning the standard sizes of common dining vessels (plates, bowls) or other visual references in the image~\cite{Sharma_2024_CVPR}.

\subsection{The Occlusion Problem}
Visual systems can only estimate the volume of visible surfaces. The bottom of the food (obscured by the plate) and internal ingredients (e.g., the filling of a burrito or layers of a lasagna) remain hidden. Deep learning models currently rely on learning visual to ``guess'' this hidden geometry, often leading to underestimation in complex, piled foods.

However, we anticipate a rise in \textit{amodal completion} networks that leverage generative diffusion models to probabilistically infer the hidden back-facing geometry of food items, effectively ``learning'' the invisible volume based on learned structural priors. Personalized approaches also use the individual's eating habit to make a more informed guess about the contents of the food rather than relying solely on visual features~\cite{pan2022personal}. 

\subsection{From Volume to Mass: The Density Gap}
Dietary assessment ultimately requires energy (kCal) estimation, not just volume ($cm^3$). The relationship between the two is governed by density ($\rho$), which varies drastically even within the same food category (e.g., a fluffy croissant vs. a dense bagel). Visual-only methods struggle to distinguish these textural nuances.

Recently, \textit{Multimodal Large Language Models (MLLMs)} offer a promising solution by combining visual features with textual context (e.g., a user's description of ``low-carb'' or ``dense''), which foundation models can use to refine density estimates~\cite{coburn2025acetada}. Additionally, accurate mapping of dietary databases to classification results (e.g. USDA's FNDDS database~\cite{montville2013usda}) can provide estimates on the density of different foods.

\section{Conclusion}

The field of image-based food portion estimation is undergoing a fundamental shift from active, hardware-dependent reconstruction to passive, AI-driven inference. Early geometric approaches, utilizing structured light and multi-view stereo, provided ground-truth reliability but failed to achieve widespread adoption due to hardware constraints and user burden. The advent of Deep Learning has enabled monocular solutions that trade a small margin of geometric accuracy for vastly improved usability.

However, true accuracy remains challenging due to inherent scale and density ambiguities. By combining visual data with semantic reasoning from large foundation models, the next generation of IBDA systems could resolve some of the issues, finally making automated, precise dietary tracking a reality for chronic disease management.

{
\bibliographystyle{IEEEtran}
\bibliography{references}
}

\end{document}